\documentclass[runningheads]{llncs}
\usepackage{graphicx}
\usepackage{amsmath}
\usepackage{amssymb}
\usepackage{algorithm}
\usepackage{algpseudocode}
\usepackage{colortbl}
\usepackage{xcolor}
\usepackage{arydshln} 
\usepackage{enumitem}
\usepackage{hyperref}

\definecolor{LightCyan}{rgb}{0.93,0.95,1.0}

\newcommand\blfootnote[1]{%
  \begingroup
  \renewcommand\thefootnote{}\footnote{#1}%
  \addtocounter{footnote}{-1}%
  \endgroup
}

\begin{document}
\title{Unsupervised Video Anomaly Detection with Diffusion Models Conditioned on Compact Motion Representations}
\titlerunning{U-VAD with Diffusion Models Conditioned on Compact Motion Rep.}
%
\author{
    Anil Osman Tur\inst{1,2}$^*$ \and
    Nicola Dall'Asen\inst{1,3}$^*$ \and
    Cigdem Beyan\inst{1} \and 
    Elisa Ricci\inst{1,2}
}
\authorrunning{Tur and Dall'Asen et al.}

\institute{University of Trento, Trento, Italy \and
Fondazione Bruno Kessler, Trento, Italy \and
University of Pisa, Pisa, Italy}
\maketitle              
\vspace{-0.5cm}
\begin{abstract}
This paper aims to address the unsupervised video anomaly detection (VAD) problem, which involves classifying each frame in a video as normal or abnormal, without any access to labels. To accomplish this, the proposed method employs conditional diffusion models, where the input data is the spatiotemporal features extracted from a pre-trained network, and the condition is the features extracted from compact motion representations that summarize a given video segment in terms of its motion and appearance. Our method utilizes a data-driven threshold and considers a high reconstruction error as an indicator of anomalous events. This study is the first to utilize compact motion representations for VAD and the experiments conducted on two large-scale VAD benchmarks demonstrate that they supply relevant information to the diffusion model, and consequently improve VAD performances \textit{w.r.t} the prior art. Importantly, our method exhibits better generalization performance across different datasets, notably outperforming both the state-of-the-art and baseline methods. The code of our method is available \href{https://github.com/AnilOsmanTur/conditioned_video_anomaly_diffusion}{HERE}\blfootnote{$^*$These authors contributed equally.}.

\end{abstract}

\section{Introduction}
\label{sec:intro}
\vspace{-0.3cm}
Detecting anomalous events in videos automatically is a crucial task of computer vision that has relevance to numerous applications, including but not limited to intelligent surveillance and activity recognition \cite{jebur2022review,sultani2018real,liu2018ano_pred,chandola2009anomaly,mohammadi2021image,beyan2013detecting}. 
Video anomaly detection (VAD) can be particularly difficult because abnormal events in the real world are infrequent and can belong to an unbounded number of categories. As a result, traditional supervised methods might not be suitable for this task since balanced normal and abnormal samples are typically unavailable for training. Moreover, VAD models are challenged by the contextual and often ambiguous nature of abnormal events, despite their sparsity and diversity \cite{ren2021deep}. As a result, VAD is commonly carried out using a \textit{one-class learning} approach, in which only normal data are provided during training~\cite{ravanbakhsh2017abnormal,smeureanu2017deep,gutowska2023constructing,zhou2019anomalynet,kim2021semi}. However, given the dynamic nature of real-world applications and the wide range of normal classes, it is not practical to have access to every type of normal training data. Therefore, when using a one-class classifier, there is a high risk of misclassifying an unseen normal event as abnormal because its representation might be significantly different from the representations learned from normal training data \cite{chandola2009anomaly}. To address the aforementioned challenge of data availability, some researchers have implemented \textit{weakly supervised} VAD that do not require per-frame annotations but instead leverage video-level labels \cite{majhi2021dam,tian2021weakly}. In weakly supervised VAD, unlike its one-class counterpart, a video is considered anomalous if even a single frame within it is labeled as anomalous. Conversely, a video is labeled as normal only when all frames within it are labeled as normal. However, such approaches lack localizing the abnormal portion of the video, which can be impractical when dealing with long videos. Also, it is important to note that labeling a video as normal still requires the inspection of entire frames \cite{zaheer2022generative}.
A more recent approach to VAD is \textbf{unsupervised learning}, in which \emph{unlabelled} videos are used as input and the model learns to classify each frame as normal or anomalous, allowing to localize the abnormal frames. Unlike a one-class classifier, unsupervised VAD does not make any assumptions about the distribution of the training data and does not use any labels during model training. However, it is undoubtedly more challenging to arrive at the performance of other VAD approaches that use labeled training data \cite{zaheer2022generative}.

This study focuses on performing unsupervised VAD in complex surveillance scenarios by relying solely on the reconstruction capability of the probabilistic generative model called \textbf{diffusion models} \cite{Karras2022edm}. The usage of generative models (e.g., autoencoders) is common for one-class VAD \cite{gong2019memorizing,nguyen2019anomaly,ren2021deep}. However, as shown in \cite{zaheer2022generative} for unsupervised VAD, the autoencoders might require an additional discriminator to be trained collaboratively to reach a desired level of performance. Instead, our study reveals that diffusion models constitute a more effective category of generative models for unsupervised VAD, displaying superior results when compared to autoencoders, and in some cases, even exceeding the performance of Collaborative Generative and Discriminative Models. Furthermore, we explore the application of \textbf{compact motion representations}, namely, \textbf{star representation} \cite{dos2020dynamic} and \textbf{dynamic images} \cite{bilen2017action} within a conditional diffusion model. This study marks the first attempt at utilizing these motion representations to address the VAD task. The experimental evaluation conducted on two large-scale datasets indicates that using the aforementioned compact motion representations as a condition of diffusion models is more beneficial for VAD. We also explore the transferability of unsupervised VAD methods by assessing their generalization performance when trained on one dataset and tested on another. When performing cross-dataset analysis, it becomes apparent that incorporating compact motion representations as the condition of diffusion models leads to vastly superior performance. This represents a crucial feature of the proposed method in comparison to both the state-of-the-art (SOTA) and baseline models, making it highly valuable for practical applications.

The main contributions can be summarized in three folds. (1)
    We propose an effective unsupervised VAD method, which uses compact motion representations as the condition of the diffusion models. We show that compact motion representations supply relevant information and further improve VAD performance.
    (2) Our method leads to enhanced generalization performance across datasets. Its transferability is notably better than the baseline methods and the SOTA.
 (3) We conduct a hyperparameter analysis for diffusion models, which yields insights into using them for VAD.

\vspace{-1em}

\section{Related Work}
\label{sec:related}
\vspace{-0.75em}
\textbf{Anomaly Detection.}
Anomaly refers to an entity that is rare and significantly deviates from normality. Automated anomaly detection models face challenges when detecting abnormal events from images or videos due to their sparsity, diversity, ambiguity, and contextual nature \cite{ren2021deep,chandola2009anomaly,zen2011earth}. 
Automated anomaly detection is a well-researched subject that encompasses various tasks, \textit{e.g.,} medical diagnosis, defect detection, animal behavior understanding, and fraud detection \cite{wolleb2022diffusion,beyan2013detecting,wang2019semi}. For a review of anomaly detection applications in different domains, interested readers can refer to the survey paper \cite{chandola2009anomaly}.
VAD, the task at hand, deals with complex surveillance scenarios. Zaheer et al. \cite{zaheer2022generative} categorized relevant methodologies into four groups: (a) fully supervised approaches requiring normal/abnormal annotations for each video frame in the training data, (b) one-class classification requiring only annotated training data for the normal class, (c) weakly supervised approaches requiring \emph{video-level} normal/abnormal annotations, and (d) unsupervised methods that do not require any annotations.

Labeling data is a costly and time-consuming task, and due to the rarity of abnormal events, it is impractical to gather all possible anomaly samples for fully-supervised learning.
Consequently, the most common approach to tackling VAD is to train a \textbf{one-class classifier} that learns from the \emph{normal} data \cite{ravanbakhsh2017abnormal,smeureanu2017deep,gutowska2023constructing,zhou2019anomalynet,kim2021semi}. Several of these approaches utilize hand-crafted features \cite{medioni2001event,piciarelli2008trajectory}, while others rely on deep features that are extracted using pre-trained models \cite{ravanbakhsh2017abnormal,smeureanu2017deep}. 
Generative models e.g., autoencoders and GANs have also been adapted for VAD \cite{gong2019memorizing,nguyen2019anomaly,ren2021deep}.
One-class classifiers often 
cannot prevent the well-reconstruction of anomalous test inputs, resulting in the misclassification of abnormal instances as normal. Moreover, an unseen normal instance could be misclassified as abnormal because its representation may differ significantly  from the representations learned from normal training data.
As evident, data collection is still a problem for the one-class approach because it is not practical to have access to every variety of normal training data~\cite{chandola2009anomaly,mohammadi2021image}. 
Therefore, some researchers \cite{majhi2021dam,tian2021weakly} have turned to \textbf{weakly supervised VAD}, which does not rely on fine-grained per-frame annotations, but instead use video-level labels. Consequently, a video is labeled as anomalous even if one frame is anomalous, and normal if all frames are normal. This setting is not optimal because labeling a video as normal requires inspecting all frames, and it cannot localize the abnormal portion.

On the other hand, VAD methods that use unlabelled training data are quite rare in the literature.
It is important to note that several one-class classifiers \cite{gutowska2023constructing,zhou2019anomalynet,kim2021semi} have been referred to as unsupervised, even though they use \emph{labeled normal} data.
\textbf{Unsupervised VAD} methods analyze unlabelled videos without prior knowledge of normal or abnormal events to classify each frame as normal or anomalous. The only published method addressing this definition is \cite{zaheer2022generative}, which presents a Generative Cooperative Learning among a generator (an autoencoder) and a discriminator (a multilayer perceptron) with a negative learning paradigm. The autoencoder reconstructs the normal and abnormal instances while the discriminator estimates the probability of being abnormal. Through negative learning, the autoencoder is constrained not to learn the reconstruction of anomalies using the pseudo-labels produced by the discriminator. That approach \cite{zaheer2022generative} follows the idea that anomalies occur less frequently than normal events, such that the generator should be able to reconstruct the abundantly available normal representations. Besides, it promotes temporal consistency while extracting relevant spatiotemporal features. Our method differs from \cite{zaheer2022generative} in that it relies solely on a generative architecture, specifically a conditional diffusion model. The baseline unconditional diffusion model, in some cases, surpasses the full model of \cite{zaheer2022generative} while in all cases it achieves better performance than the autoencoder of \cite{zaheer2022generative}. On the other hand, the proposed method improves the achievements of the unconditional diffusion model thanks to using compact motion representations, and importantly, it presents the best generalization results across datasets.

\noindent \textbf{Diffusion Models.} They are a family of probabilistic generative models that progressively destruct data by injecting noise, then learn to reverse this process for sample generation.~\cite{ho2020denoising,Karras2022edm}.
Diffusion models have emerged as a powerful new family of deep generative models with SOTA performance in many applications, including image synthesis, video generation, and discriminative tasks like object detection and semantic segmentation~\cite{rombach2022high,blattmann2023align}. Given that diffusion models have emerged as SOTA generative models for various tasks, we are motivated to explore their potential for VAD through our proposed method.

\noindent \textbf{Star Representation \cite{dos2020dynamic}.}
It aims to represent temporal information existing in a video in a way that the channels of output single RGB image convey the summarized time information by associating the color channels with simplified consecutive moments of the video clip. Such a representation is suitable to be the input of any CNN model and so far in the literature, it was used for
dynamic gesture recognition \cite{barros2014real,dos2020dynamic}, while this is the first time it is being used for VAD.

\noindent \textbf{Dynamic Image \cite{bilen2017action}.}
It refers to a representation of an input video sequence that summarizes the appearances of objects and their corresponding motions over time by encoding the temporal ordering of the pixels from frame to frame. This can be seen as an early fusion technique since the frames are combined into a single representation before further processing them such as with. It has been used for action and gesture recognition \cite{bilen2017action,wang2017ordered} and visual activity modeling \cite{beyan2019personality,shahids}, however, it has never been used for VAD.

\begin{figure}[!ht]
\vspace{-0.3cm}
\centering
\includegraphics[width=0.7\textwidth]{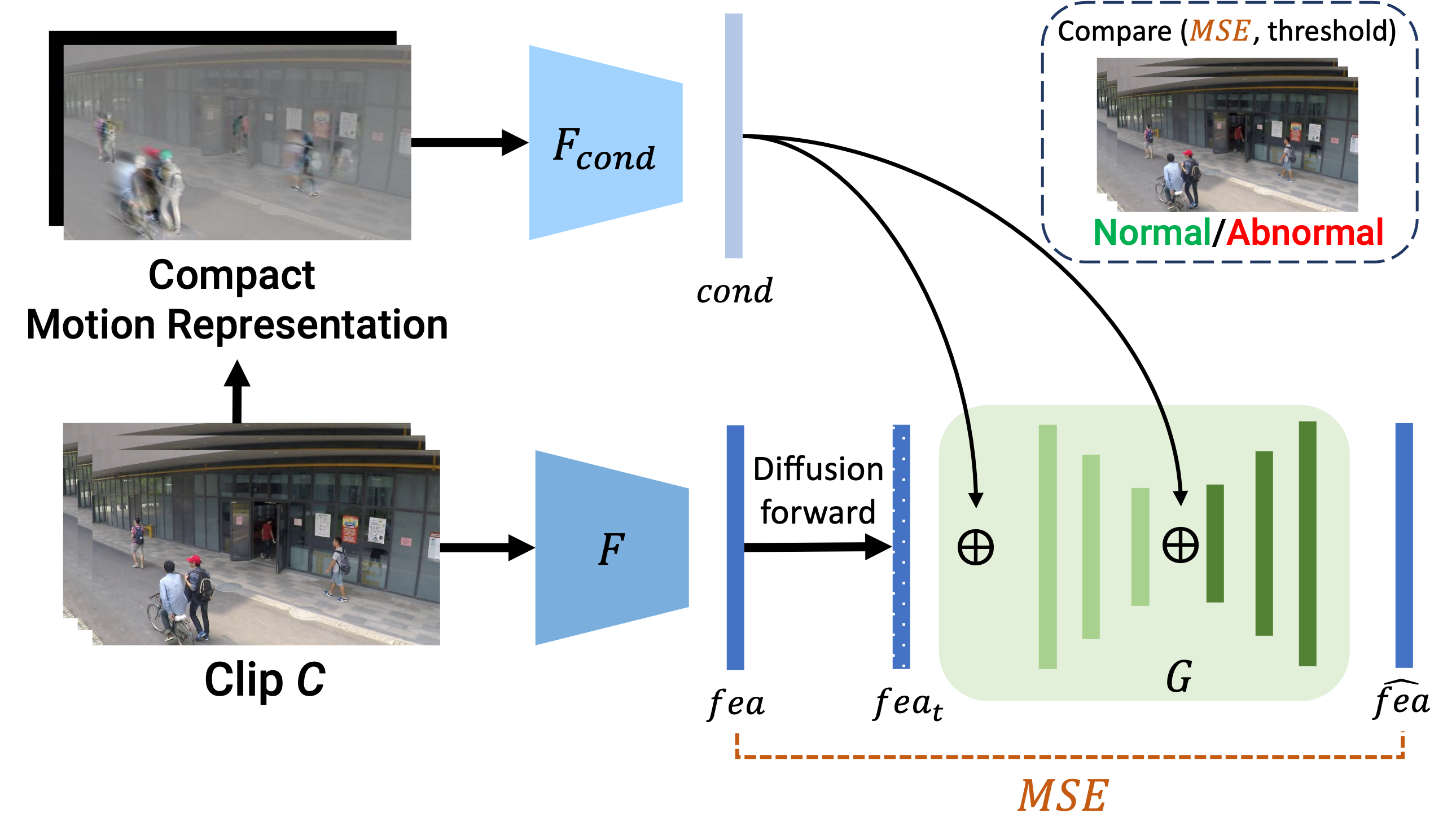}
\vspace{-0.3cm}
\caption{An illustration of the proposed method. For definitions of the abbreviations used, please refer to the text.} \label{fig:proposed}
\vspace{-0.5cm}
\end{figure}

\section{Proposed Method}
\label{sec_method}
\vspace{-0.3cm}
We design a method to use diffusion models to tackle the unsupervised VAD, \textit{i.e.} to classify each frame in a video as normal or abnormal without using the labels. To provide a frame-based prediction, we classify a video clip of consecutive $N$ frames and then slide this window along the video. We build our model on top of diffusion models, in particular, \textit{k-diffusion}~\cite{Karras2022edm}, which has shown better performance \textit{w.r.t} DDPM~\cite{ho2020denoising}. 
To overcome the heavy computational burden of dealing with video clips, we operate in the latent space of a pre-trained network that extracts clip-level features. We then leverage the generative capabilities of diffusion models to reconstruct noised clip features and, based on the reconstruction error, decide whether the clip is normal or abnormal with a data-driven threshold. While this formulation leads to SOTA performance, we further condition the diffusion process with compact motion information coming from the video clip (see Sec.~\ref{sec:compactMotion}) to better guide the reverse process and achieve better performance. An overview of our method is provided in Fig.~\ref{fig:proposed}.
\vspace{-0.5em}
\subsection{Diffusion Model}
\label{sec:diff}
\vspace{-0.2cm}
Diffusion models apply a progressive addition of Gaussian noise $\epsilon_t$ of standard deviation $\sigma_t$ to an input data point $x_T$ sampled from a distribution $p_{data}(x)$ for each timestep $t \in [0, T]$. The noised distribution $p(x,\sigma)$ becomes isotropic Gaussian and allows efficient sampling of new data points $x_0 \sim \mathcal{N}(0, \sigma^2_{max}\mathbf{I})$. These data are gradually denoised with noise levels $\sigma_0 = \sigma_{max} > \sigma_{1} > \dots > \sigma_{T-1} > \sigma_T = 0$ into new samples. Diffusion models are trained by minimizing the expected $L_2$ error between predicted and ground truth added noise~\cite{ho2020denoising}, \textit{i.e.}: $\mathcal{L}_{simple} = \| \epsilon_t - \hat{\epsilon} \|_2$. In this work, we use the diffusion formulation of~\cite{Karras2022edm}, which allows the network to perform either $\epsilon$ or $x_0$ prediction, or something in between, depending on the noise scale $\sigma_t$, nullify the error amplification that happens in DDPM~\cite{Karras2022edm}. The denoising network $D_\theta$ formulation as follows:
\begin{equation}
\vspace{-0.75em}
D_\theta(x; \sigma_t) = c_{skip}(\sigma_t) ~x + c_{out}(\sigma_t) ~G_\theta \big( c_{in}(\sigma_t) ~x; ~c_{noise}(\sigma-T) \big),
\label{eq:preconditioning}
\end{equation}
where $G_\theta$ becomes the effective network to train, $c_{skip}$ modulates the skip connection, $c_{in}(\cdot)$ and $c_{out}(\cdot)$ scale input and output magnitudes, and $c_{noise}(\cdot)$ scales $\sigma$ to become suitable as input for $F_\theta$. Formally, given a video clip $C$ of $N$ frames, \textit{i.e.} $C \in \mathbb{R}^{N\times 3\times H\times W}$, we first extract features from a pre-trained 3D-CNN $\mathcal{F}$ to obtain a feature vector $fea \in \mathbb{R}^{f}$, with $f$ the latent dimension of the network. We then use this latent representation in the diffusion process to reconstruct them without using any label.  

We leverage the fact that denoising does not necessarily have to start from noise with variance $\sigma^2_{max}$, but it can place at any arbitrary timestep $t \in (0, T]$, as shown in~\cite{meng2021sdedit}. We can therefore sample $fea_t \sim \mathcal{N}(fea, \sigma^2_t)$ and run the diffusion reverse process on it to reconstruct $fea_T$. The choice of $t$ allows balancing the amount of information destroyed in the forward process, and we exploit this fact to remove the frequency components associated with anomalies. We then measure the reconstruction goodness in terms of $MSE$, with a higher reconstruction error \textit{possibly} indicating that the clip is anomalous. When deciding whether a video frame is anomalous or not, we adopt the data-driven thresholding mechanism of \cite{zaheer2022generative}. The decision for a single video frame is made by keeping the distribution of the reconstruction loss ($MSE$) of each clip over a \emph{batch}. The feature vectors resulting in higher reconstruction error refer to anomalous clips and vice versa. This decision is made through the data-driven threshold $L_{th}$, defined as $L_{th}$ $=$ $\mu_p$ $+$ $k$ $\sigma_p$ where $k$ is a constant, $\mu_p$ and $\sigma_p$ are the mean and standard deviation of the reconstruction error for each batch.
\begin{figure}[!t]
\vspace{-2.5em}
\centering
\includegraphics[width=0.75\textwidth]{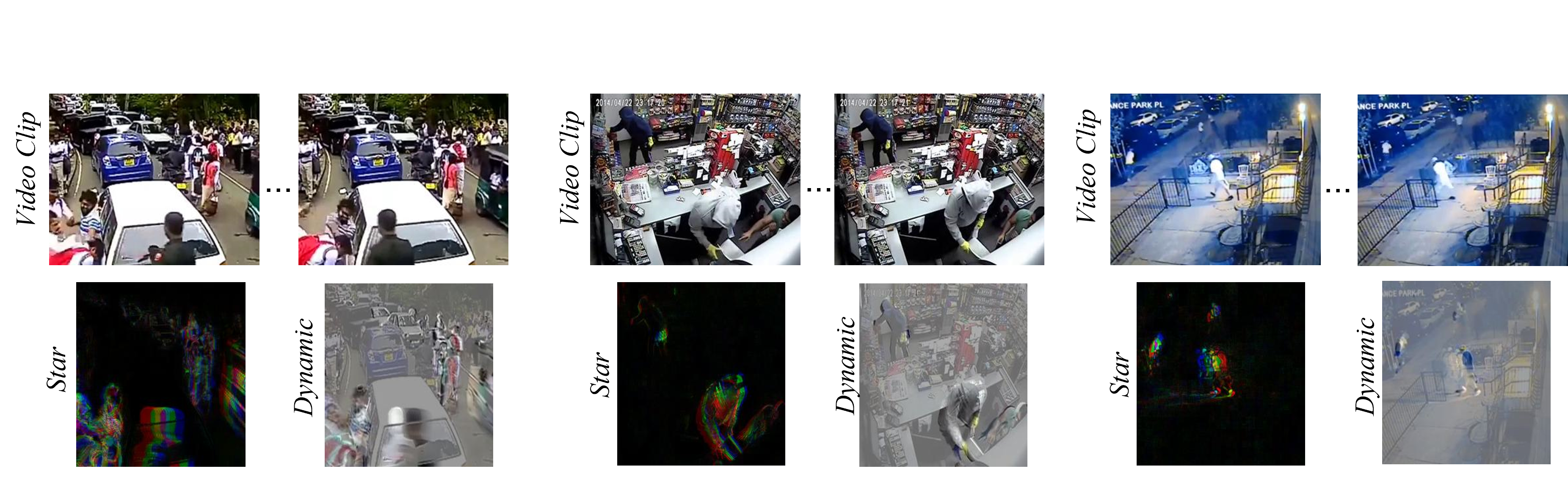}
\vspace{-0.3cm}
\caption{Examples of star representation and dynamic image for a given video clip.}
\label{fig:motion}
\vspace{-1.5em}
\end{figure}

\subsection{Compact Motion Representations}
\label{sec:compactMotion}
\vspace{-0.2cm}
We further extend the described diffusion model to incorporate compact motion representation in the process to provide rich motion information. We compute this representation using two different approaches: Star representation~\cite{dos2020dynamic} or Dynamic Image~\cite{bilen2017action}.
Visual examples of these two representations are presented in Fig.~\ref{fig:motion} and a complete description is presented as follows. 

\noindent \textbf{Star RGB Images.}
The objective of using star representation is to depict the time-based data present in an input RGB video \cite{barros2014real,dos2020dynamic}.
The star representation matrix $M$ computation is computed as given in Eq. \ref{eq:motionImage} where $I_k(i,j)$ represents the RGB vectors of a pixel at a given $(i,j)$ position at $k-th$ frame and $\lambda$ is the cosine similarity of the RGB vectors.
By using such a cosine similarity star representation also includes the information change in hue and saturation.
\vspace{-0.2cm}
\begin{equation}
\small{ M(i,j) = \sum_{k=2}^{N}\left(1 - \frac{\lambda}{2} \right). |\parallel I_{k-1}(i,j)\parallel_2 - \parallel I_{k-1}(i,j)\parallel_2|, }
\label{eq:motionImage}
\vspace{-0.5em}
\end{equation}
where $N$ is the length of the video clip.
To create an RGB image as the output, each video segment is divided equally into three sub-videos such that each sub-video is used for generating one of the RGB channels. Thus, the resulting image channels convey the summarized information of consecutive moments. 

\noindent \textbf{Dynamic Image Computation.}
A dynamic image presents a summary of object appearances and their motions throughout an input video sequence by encoding the sequential order of pixels from one frame to another. 
Dynamic image computation uses RGB images directly by multiplying the video frames by $\alpha_t$ coefficient and summing them to generate the output image with the formula given $d^* = \sum_{k=1}^{N} \alpha_k I_k , ~ \alpha_k = 2k - N - 1$,
where $I_k$ is the $kth$ image of the video segment and $N$ is the number of frames in the video segment.

\noindent \textbf{Conditioning on Compact Motion Representation.}
After extracting the compact motion representation of a clip $C$, we obtain the conditioning feature vector $cond$ through a pre-trained 2D-CNN $F_{cond}$. We inject this both in the encoder and in the decoder part of our network $G$ by summing with the input features. To deal with the different dimensionality of the two blocks, we use 2 linear projections to obtain vectors of the same size as the input.

\vspace{-0.75em}
\section{Experimental Analysis and Results}
\label{sec:exp}
\vspace{-0.75em}
The \textbf{evaluation metric} employed in this study is the Area Under the Receiver Operating Characteristic (ROC) Curve (AUC), which is determined using frame-level annotations of the test videos within the datasets, consistent with established VAD methodologies. In order to evaluate and compare the effectiveness of the proposed approach, the experiments were carried out on two mainstream large-scale unconstrained \textbf{datasets}: UCF-Crime \cite{sultani2018real} and ShanghaiTech \cite{liu2018ano_pred}.
The \textbf{UCF-Crime dataset} \cite{sultani2018real} was obtained from diverse CCTV cameras that possess varying field-of-views. It consists of a total of 128 hours of videos, with annotations for 13 distinct anomalous events e.g., road accidents, theft, and explosions. To ensure fair comparisons with the SOTA, we utilized the standardized training and testing splits of the dataset, which consist of 810 abnormal and 800 normal videos for training, and 130 abnormal and 150 normal videos for testing, without utilizing the labels. On the other hand, the \textbf{ShanghaiTech dataset} \cite{liu2018ano_pred} was recorded using 13 distinct camera angles under challenging lighting conditions. For our study, we utilized the training split, which comprises 63 abnormal and 174 normal videos, as well as the testing split, consisting of 44 abnormal and 154 normal videos, in accordance with SOTA conventions. 
\vspace{-0.75em}
\subsection{Implementation Details}
\label{subsec:ImpDet}
\vspace{-0.5em}
\noindent\textbf{Architecture.} In line with \cite{zaheer2022generative}, we use 16 non-overlapping frames to define a video clip, and we use pre-trained 3D-ResNext101 or 3D-ResNet18 as feature extractor $F$~\cite{chandola2009anomaly,zaheer2022generative}. After computing the compact motion representation, we extract a single conditioning vector with $F_{cond}$ with a pre-trained ResNet50 or ResNet18 due to their widespread use together with such motion representations \cite{beyan2019personality,shahids}. We use an MLP with an encoder-decoder structure as the denoising network $G$, and the encoder is comprised of three layers with sizes of \{1024, 512, 256\}, while the decoder has hidden dimensions of \{256, 512, 1024\}. The timestep information $\sigma_t$ is transformed via Fourier embedding and integrated into the network by FiLM layers \cite{perez2018film}, while the conditioning on compact motion representation is applied after timestep integration by summation to the inputs.

\noindent\textbf{Training and sampling.} The learning rate scheduler and EMA of the model are set to the default values of \textit{k-diffusion}, which include an initial learning rate of $2\times10^{-4}$ and InverseLR scheduling. The weight decay is set at $1\times10^{-4}$. Training is conducted for 30 epochs with a batch size of 256, while testing is performed on 8192 samples as in previous literature~\cite{zaheer2022generative}. Several hyperparameters affect the diffusion process in \textit{k-diffusion}, and given the novelty of the task at hand, we do not rely on parameters from prior literature. We, therefore, conduct an extensive exploration of the effects of training and testing noise. Training noise is distributed according to a log-normal distribution with parameters $(P_{mean}, P_{std})$, while sampling noise is controlled by $\sigma_{min}$ and $\sigma_{max}$, and below, we investigate their role. For the diffusion reverse process, we use LMS sampler with the number of steps $T$ set to 10.

\begin{table*}[!t]
\caption{Performance comparisons with the SOTA and the baseline methods on ShanghaiTech \cite{liu2018ano_pred} dataset. The best results are in bold. The second best results are \underline{underlined}. The full model of \cite{zaheer2022generative} includes generator, negative learning, and discriminator. $NA$ stands for not-applicable. Results with $\diamond$ are taken from \cite{zaheer2022generative}.}
\label{table:compare_shang}
\centering
\resizebox{0.7\linewidth}{!}{
\begin{tabular}{lccc}
\hline
Method & Feature & Condition & AUC (\%) \\
\hline
\multicolumn{4}{c}{State-of-the-art Methods} \\ \hline
Kim et al. \cite{kim2021semi}$^\diamond$ & 3D-ResNext101 & NA & 56.47 \\
Autoencoder \cite{zaheer2022generative} & 3D-ResNext101 & NA & 62.73  \\
Autoencoder \cite{zaheer2022generative} & 3D-ResNet18 & NA & 69.02 \\
Full model \cite{zaheer2022generative} & 3D-ResNext101 & NA & 72.41 \\
Full model \cite{zaheer2022generative} & 3D-ResNet18 & NA & 71.20 \\ \hline
\multicolumn{4}{c}{Baseline Methods} \\ \hline
Diffusion & 3D-ResNext101  & - & 68.88 \\
Diffusion & 3D-ResNet18 & - & 76.10 \\ 
Diffusion & Star Rep. \cite{dos2020dynamic} w/ ResNet18 & - & 62.81 \\
Diffusion & Star Rep. \cite{dos2020dynamic} w/ ResNet50  & - & 59.55 \\
Diffusion & Dyn. Img. \cite{bilen2017action} w/ ResNet18  & - & 62.88 \\
Diffusion & Dyn. Img. \cite{bilen2017action} w/ ResNet50  & - & 64.96 \\ \hline
\multicolumn{4}{c}{Other Conditional Diffusion Models} \\ \hline
Diffusion &  Star Rep. \cite{dos2020dynamic} w/ ResNet18 & 3D-ResNext101 & 64.87 \\ 
Diffusion &  Star Rep. \cite{dos2020dynamic} w/ ResNet50 & 3D-ResNext101 & 65.01 \\
Diffusion &  Star Rep. \cite{dos2020dynamic} w/ ResNet18 & 3D-ResNext18 & 64.03 \\
Diffusion &  Star Rep. \cite{dos2020dynamic} w/ ResNet50 & 3D-ResNext18 & 64.15 \\
Diffusion &  Dyn. Img. \cite{bilen2017action} w/ ResNet18 & 3D-ResNext101 & 66.66 \\
Diffusion &  Dyn. Img. \cite{bilen2017action} w/ ResNet50 & 3D-ResNext101 & 64.24 \\
Diffusion &  Dyn. Img. \cite{bilen2017action} w/ ResNet18 & 3D-ResNext18 & 65.02 \\
Diffusion &  Dyn. Img. \cite{bilen2017action} w/ ResNet50 & 3D-ResNext18 & 65.26 \\ \hline
\multicolumn{4}{c}{Proposed Method} \\ \hline
\rowcolor{LightCyan} Diffusion & 3D-ResNext 101  & Star Rep. \cite{dos2020dynamic} w/ ResNet18 & 65.12 \\
\rowcolor{LightCyan} Diffusion & 3D-ResNext 101  & Star Rep. \cite{dos2020dynamic} w/ ResNet50 & 65.17 \\
\rowcolor{LightCyan} Diffusion & 3D-ResNext 101  & Dyn. Img. \cite{bilen2017action} w/ ResNet18 & 66.36 \\
\rowcolor{LightCyan} Diffusion & 3D-ResNext 101  & Dyn. Img. \cite{bilen2017action} w/ ResNet50 & 65.09 \\
\rowcolor{LightCyan} Diffusion & 3D-ResNet18  & Star Rep. \cite{dos2020dynamic} w/ ResNet18 & 76.36 \\
\rowcolor{LightCyan} Diffusion & 3D-ResNet18  & Star Rep. \cite{dos2020dynamic} w/ ResNet50 & \textbf{77.18} \\
\rowcolor{LightCyan} Diffusion & 3D-ResNet18  & Dyn. Img. \cite{bilen2017action} w/ ResNet18 & 74.61 \\
\rowcolor{LightCyan} Diffusion & 3D-ResNet18  & Dyn. Img. \cite{bilen2017action} w/ ResNet50 & \underline{76.16} \\
\hline
\end{tabular}}
\vspace{-0.5cm}
\end{table*}

\vspace{-1em}
\subsection{Results}
\label{sec:results}
\vspace{-0.75em}
We first compare our method's results with SOTA and baseline methods. 
Then, we report the results of the cross-dataset evaluation, where the training and validation sets are from a different domain than the test split. Finally, we analyze how the hyperparameters of the diffusion models affect VAD performance.

\begin{table*}[!t]
\caption{Performance comparisons with the SOTA and the baseline methods on UCF-Crime \cite{sultani2018real} dataset. The best results are in bold. The second best results are \underline{underlined}. The full model of \cite{zaheer2022generative} includes generator, negative learning, and discriminator. $NA$ stands for not-applicable. Results with $\diamond$ are taken from \cite{zaheer2022generative}.}
\label{table:compare_UCFC}
\centering
\resizebox{0.68\linewidth}{!}{
\begin{tabular}{lccc}
\hline
 Method & Feature & Condition & AUC (\%) \\
\hline
\multicolumn{4}{c}{State-of-the-art Methods} \\ \hline
Kim et al. \cite{kim2021semi}$^\diamond$ & 3D-ResNext101 & NA & 52.00 \\
Autoencoder \cite{zaheer2022generative} & 3D-ResNext101 & NA & 56.32  \\
Autoencoder \cite{zaheer2022generative} & 3D-ResNet18 & NA & 49.78 \\
Full model \cite{zaheer2022generative} & 3D-ResNext101 & NA & \textbf{68.17} \\
Full model \cite{zaheer2022generative} & 3D-ResNet18 & NA & 56.86 \\ \hline
\multicolumn{4}{c}{Baseline Methods} \\ \hline
Diffusion & 3D-ResNext101  & - & 62.91 \\
Diffusion & 3D-ResNet18 & - & 65.22 \\ 
Diffusion & Star Rep. \cite{dos2020dynamic} w/ ResNet18 & - & 59.60 \\
Diffusion & Star Rep. \cite{dos2020dynamic} w/ ResNet50  & - & 61.14 \\
Diffusion & Dyn. Img. \cite{bilen2017action} w/ ResNet18  & - & 60.14 \\
Diffusion & Dyn. Img. \cite{bilen2017action} w/ ResNet50  & - & 62.73 \\ 
\hline
\multicolumn{4}{c}{Other Conditional Diffusion Models} \\ \hline
Diffusion &  Star Rep. \cite{dos2020dynamic} w/ ResNet18 & 3D-ResNext101 & 59.26 \\
Diffusion &  Star Rep. \cite{dos2020dynamic} w/ ResNet50 & 3D-ResNext101 & 63.20 \\
Diffusion &  Star Rep. \cite{dos2020dynamic} w/ ResNet18 & 3D-ResNext18 & 61.14 \\
Diffusion &  Star Rep. \cite{dos2020dynamic} w/ ResNet50 & 3D-ResNext18 & 60.78 \\
Diffusion &  Dyn. Img. \cite{bilen2017action} w/ ResNet18 & 3D-ResNext101 & 58.23 \\
Diffusion &  Dyn. Img. \cite{bilen2017action} w/ ResNet50 & 3D-ResNext101 & 61.04 \\
Diffusion &  Dyn. Img. \cite{bilen2017action} w/ ResNet18 & 3D-ResNext18 & 65.06 \\
Diffusion &  Dyn. Img. \cite{bilen2017action} w/ ResNet50 & 3D-ResNext18 & 61.27 \\ \hline 
\multicolumn{4}{c}{Proposed Method} \\ \hline
\rowcolor{LightCyan} Diffusion & 3D-ResNext101  & Star Rep. \cite{dos2020dynamic} w/ ResNet18 & 58.82 \\ 
\rowcolor{LightCyan} Diffusion & 3D-ResNext101  & Star Rep. \cite{dos2020dynamic} w/ ResNet50 & 63.00 \\
\rowcolor{LightCyan} Diffusion & 3D-ResNext101  & Dyn. Img. \cite{bilen2017action} w/ ResNet18 & 60.12 \\
\rowcolor{LightCyan} Diffusion & 3D-ResNext101  & Dyn. Img. \cite{bilen2017action} w/ ResNet50 & 63.52 \\
\rowcolor{LightCyan} Diffusion & 3D-ResNet18  & Star Rep. \cite{dos2020dynamic} w/ ResNet18 & 63.67 \\
\rowcolor{LightCyan} Diffusion & 3D-ResNet18  & Star Rep. \cite{dos2020dynamic} w/ ResNet50 & \underline{66.85} \\
\rowcolor{LightCyan} Diffusion & 3D-ResNet18  & Dyn. Img. \cite{bilen2017action} w/ ResNet18 & 60.69 \\
\rowcolor{LightCyan} Diffusion & 3D-ResNet18  & Dyn. Img. \cite{bilen2017action} w/ ResNet50 & 66.11 \\ 
\hline
\end{tabular}}
\vspace{-2.65em}
\end{table*}

\noindent \textbf{Performance Comparisons.}
The performance of the proposed method together with the SOTA and baseline methods' (i.e., unconditional diffusion model) results are given in Tables \ref{table:compare_shang} and \ref{table:compare_UCFC} for the ShanghaiTech \cite{liu2018ano_pred} and UCF-Crime \cite{sultani2018real} datasets, respectively. These tables also include an ablation study such that the condition of the diffusion models is changed between star representation, dynamic images, and spatiotemporal features, in addition to changing the feature backbone between 3D-ResNext101 and 3D-ResNet18, and the motion representation backbone between ResNet50 and ResNet18. 

As seen in Table \ref{table:compare_shang}, the proposed method outperforms all others on the ShanghaiTech \cite{liu2018ano_pred} dataset, achieving the best results by surpassing the SOTA autoencoder \cite{zaheer2022generative} by 14.45\%, the SOTA collaborative generative and discriminative model \cite{zaheer2022generative} by 4.77\%, and the SOTA \cite{kim2021semi} by 20.71\%. The proposed method also improves upon the unconditional diffusion models (i.e., baselines) by 1.08\%. It is worth noting that other conditional diffusion models, i.e., using spatiotemporal features as the condition and compact motion representation as input, are occasionally less effective than our method, with the proposed method surpassing them by 10.52\%. The best performance is achieved by using a 3D-ResNet18 as the feature backbone, star representation as the condition, and ResNet50 as the corresponding backbone. On the other hand, for the UCFC dataset \cite{sultani2018real} (Table \ref{table:compare_UCFC}), the proposed method achieves the second-highest score after the more complex model of \cite{zaheer2022generative}, which employs a generator, discriminator and negative learning. Nonetheless, our method outperforms the SOTA autoencoder \cite{zaheer2022generative} by 10.53\% and the SOTA \cite{kim2021semi} by 14.85\%. It also demonstrates superior performance compared to the baseline and the other conditional diffusion models by 1.63\% and 1.79\%, respectively. Furthermore, the optimal performance of the proposed method for this dataset is achieved by utilizing 3D-ResNet18 as the feature backbone, star representation as the condition, and ResNet50 as the condition backbone. 

\begin{table}[!t]
\caption{Cross-dataset analysis (Training dataset -$>$ Testing dataset). The best results are in bold. The second best results are \underline{underlined}. The full model of \cite{zaheer2022generative} includes generator, negative learning, and discriminator. $NA$ stands for not-applicable.}
\label{table:compare}
\centering
\resizebox{0.68\linewidth}{!}{
\begin{tabular}{lccc}
\hline
Method & Feature & Condition & AUC (\%) \\
\hline
\multicolumn{4}{c}{UCFC -$>$ ShanghaiTech} \\ \hline
Autoencoder \cite{zaheer2022generative} & 3D-ResNext101 & NA & 55.86 \\
Autoencoder \cite{zaheer2022generative} & 3D-ResNet18 & NA & 47.48 \\
Full model \cite{zaheer2022generative} & 3D-ResNext101 & NA & 55.94 \\
Full model \cite{zaheer2022generative} & 3D-ResNet18 & NA & 49.19 \\
Diffusion (Baseline) & 3D-ResNet18 & - & 60.55 \\ 
Diffusion (Baseline) & Star Rep. \cite{dos2020dynamic} w/ ResNet50  & - & 54.67 \\
Diffusion (Baseline) & Dyn. Img. \cite{bilen2017action} w/ ResNet50  & - & 58.14 \\
\rowcolor{LightCyan} Diffusion (Proposed) & 3D-ResNet18  & Star Rep. \cite{dos2020dynamic} w/ ResNet50 & \textbf{64.54} \\
\rowcolor{LightCyan} Diffusion (Proposed) & 3D-ResNet18  & Dyn. Img. \cite{bilen2017action} w/ ResNet50 & \underline{63.58} \\
\hline
\multicolumn{4}{c}{ShanghaiTech -$>$ UCFC} \\ \hline
Autoencoder \cite{zaheer2022generative} & 3D-ResNext101 & NA & 52.45 \\
Autoencoder \cite{zaheer2022generative} & 3D-ResNet18 & NA & 46.53 \\
Full model \cite{zaheer2022generative} & 3D-ResNext101 & NA & 52.29 \\
Full model \cite{zaheer2022generative} & 3D-ResNet18 & NA & 49.57 \\
Diffusion (Baseline) & 3D-ResNet18 & - & 63.97 \\
Diffusion (Baseline) & Star Rep. \cite{dos2020dynamic} w/ ResNet50  & - & 60.21 \\
Diffusion (Baseline) & Dyn. Img. \cite{bilen2017action} w/ ResNet50  & - & 60.75 \\
\rowcolor{LightCyan} Diffusion (Proposed) & 3D-ResNet18  & Star Rep. \cite{dos2020dynamic} w/ ResNet50 & \textbf{65.17} \\
\rowcolor{LightCyan} Diffusion (Proposed) & 3D-ResNet18  & Dyn. Img. \cite{bilen2017action} w/ ResNet50 & \underline{64.97} \\
\hline
\end{tabular}}
\vspace{-0.5em}
\end{table}

\noindent \textbf{Cross-dataset Analysis.}
When performing this analysis, we take into consideration the results presented in Tables \ref{table:compare_shang} and \ref{table:compare_UCFC} such that we select the combinations of input feature and condition backbone that yield the best results. Table \ref{table:compare} shows that our methods achieve significantly better results in cross-dataset analysis, regardless of which compact motion representation is used as the condition, compared to all other baselines and SOTA methods. Notably, the performance of the proposed method is remarkable (8.6-17.06\% better) in comparison to both the generative model and full model proposed by \cite{zaheer2022generative}. On the other hand, the baseline unconditional diffusion model that utilizes spatiotemporal features outperforms the baseline unconditional diffusion model that uses compact motion representations. The relative effectiveness of the proposed method is of significant practical importance, as in most cases, the deployment domain differs from the domain on which the model is trained. 

\noindent \textbf{Hyperparameter Analysis.}
We study the effect of the \textbf{training noise} on the learning process, and we find that baseline diffusion and our method both achieve higher results with smaller values of noise, meaning a lower $P_{mean}$. Importantly, our method generally achieves better performance than the baseline, given the same parameters, for a wider choice range of training noise parameters, making it less sensitive to this choice. We explore the effect of $P_{mean} \in [-5, -0.5]$ and $P_{std} \in [0.5, 2.]$. 
On the other hand, recalling that $t$ closer to zero indicates a point closer to an isotropic Gaussian distribution, we explore the \textbf{effect of} different $t$ as \textbf{the starting point of the reverse process}. While the baseline unconditional diffusion achieves its best performance with $t=4$ and $t=6$, we find that our method achieves better performance in high-noise areas ($t=1, t=2$), effectively allowing the removal of more information from the clip vector, and proving the effectiveness of conditioning on motion representation for the task at hand.

\vspace{-0.3cm}
\section{Conclusions}
\label{sec:conc}
\vspace{-0.5em}
We have presented a novel approach for unsupervised VAD, which can accurately identify and locate anomalous frames by utilizing only the reconstruction capabilities of diffusion models. Our conditional diffusion model uses features extracted from compact motion representations as the condition while it takes the spatiotemporal features extracted from pre-trained networks as the input. By doing so, we show the contribution of the compact motion representations, \textit{i.e.,} our method succeeded in improving the SOTA VAD results while also demonstrating remarkable transferability across domains.
Note that the unsupervised nature of our approach allows for an anomaly detection system to begin identifying abnormalities based solely on observed data, without any human intervention. If no abnormal events have occurred, the system may mistakenly identify rare normal events as abnormal. However, it is expected that such anomaly systems operate for a longer period of time, thus, the likelihood of having no abnormal events decreases significantly. In the future, we aim to modify our method in a way that it can operate on edge devices with near real-time capabilities. 
\vspace{-1em}

\section*{Acknowledgment}
The project is partially funded by the European Union (EU) under NextGenerationEU. We acknowledge the support of the MUR PNRR project FAIR - Future AI Research (PE00000013) funded by the NextGenerationEU. E.R. is partially supported by the PRECRISIS, funded by the EU Internal Security Fund (ISFP-2022-TFI-AG-PROTECT-02-101100539). Views and opinions expressed are however those
of the author(s) only and do not necessarily reflect those of the EU or The European Research
Executive Agency. Neither the EU nor the granting authority can be held responsible for
them. The work was
carried out in the Vision and Learning joint laboratory of FBK and UNITN.

%
%
%

\bibliographystyle{splncs04}
\bibliography{bibliography}

\end{document}